\documentclass[11pt,a4paper]{article}
\usepackage[hyperref]{acl2020}
\usepackage{times}
\usepackage{latexsym}

\usepackage{helvet} 
\usepackage{courier}  
\usepackage{url}  
\usepackage{mathrsfs}
\usepackage{amsmath}
\usepackage{amsfonts}
\usepackage{graphicx, subfigure, float} 
\usepackage{microtype}

\aclfinalcopy  

\title{Transformation of Dense and Sparse Text Representations}

\author{
Wenpeng Hu$^{1,3,}$\thanks{\ \ Equal Contribution.}, Mengyu Wang$^{2,3,*}$, Bing Liu$^{3,}$\thanks{\ \ Corresponding Author}\\
\bf Feng Ji$^4$, Haiqing Chen$^4$, Dongyan Zhao$^{3}$, Jinwen Ma$^1$ \and Rui Yan$^{3,\dag}$\\
$^1$Department of Information Science, School of Mathematical Sciences, Peking University\\
$^2$Yuanpei College, Peking University\\
$^3$Wangxuan Institute of Computer Technology, Peking University\\
$^4$Alibaba Group\\
\{wenpeng.hu,wangmengyu,dcsliub,zhaody,ruiyan\}@pku.edu.cn, jwma@math.pku.edu.cn
}

\date{}

\begin{document}
\maketitle
\begin{abstract}
Sparsity is regarded as a desirable property of representations, especially in terms of explanation. However, its usage has been limited due to the gap with dense representations. Most NLP research progresses in recent years are based on dense representations. Thus the desirable property of sparsity cannot be leveraged. Inspired by Fourier Transformation, in this paper, we propose a novel Semantic Transformation method to bridge the dense and sparse spaces, which can facilitate the NLP research to shift from dense space to sparse space or to jointly use both spaces. The key idea of the proposed approach is to use a Forward Transformation to transform dense representations to sparse representations. Then some useful operations in the sparse space can be performed over the sparse representations, and the sparse representations can be used directly to perform downstream tasks such as text classification and natural language inference. Then, a Backward Transformation can also be carried out to transform those processed sparse representations to dense representations. Experiments using classification tasks and natural language inference task show that the proposed Semantic Transformation is effective. 
\end{abstract}

\section{Introduction}
Many studies have shown that sparsity is a desirable property of representations, especially in terms of explanation \cite{fyshe2014interpretable,faruqui2015non}. In this sense, sparse representation may hold the key to solving the explainability problem of deep neural networks. Apart from the interpretability property, sparse representation can also improve the usability of word vectors as features \cite{guo2014revisiting,chang2018xsense}. Several tasks have benefited from sparse representations, e.g., part-of-speech tagging \cite{ganchev2009posterior}, dependency parsing \cite{martins2011structured}, and supervised classification \cite{yogatama2014linguistic}.

However, much of the research advances so far for NLP tasks are based on dense representations, e.g., text classification \cite{kim2014convolutional,tang2015joint,wu2017sentence,wang2018sentiment}, natural language inference \cite{liu2019multi,kim2019semantic}, machine translation \cite{cheng2019semi,he2016dual} and generation \cite{serban2017multiresolution,zhang2019consistent,zhu-etal-2018-msmo}. The study of sparse representations is still limited. 

There are two key limitations in the existing studies of sparse representations. First, there is no study that has been done to connect dense and sparse spaces well, which makes the two types of representations relatively independent and cannot reinforce each other to achieve synergy. Second, limited work has been done to generate representations of sentences or phrases in the sparse space using sparse word embeddings. 
\begin{figure}[t]
\centering 
\includegraphics[width=0.45\columnwidth]{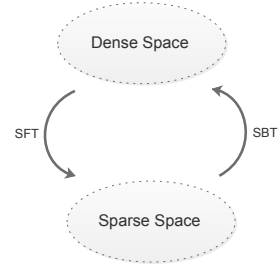}
\caption{Transformations between dense and sparse spaces. SFT and SBT denote the forward and backward transformation respectively.} 
\label{fig:intro} 
\end{figure}

Inspired by Fourier Transformation, as shown in Figure \ref{fig:intro}, this paper proposes a novel method called Semantic Transformation (ST) to address the problems. With the help of ST, dense and sparse spaces can connect with each other and will not be isolated anymore. The proposed transformation consists of two key components, namely,  Semantic Forward Transformation (SFT) and Semantic Backward Transformation (SBT) (see Section \ref{sec:ST}). SFT is designed to transform a dense representation to a sparse representation. That is, we can transform any learned dense features to sparse representations and endow the model properties that sparsity possesses. Sparse representations can also be transformed back to dense representations through SBT, before that, we can perform different operations in the sparse space to achieve different goals.

Another key innovation of this paper is that it proposes a new approach for achieving sparseness. Conventionally, penalties are commonly used to achieve sparseness~\cite{sun2016sparse,ng2011sparse,subramanian2018spine}. However, they suffer from the problems of initialization sensitivity and uncontrollable optimization. In this paper, we propose to achieve sparseness through a novel activation function, which gives an effective solution (see Section \ref{sec:SFT}). Experimental results show that the proposed activation function works very well.

In this paper, we also explore a combination method to combine words representations into sentence representations in the sparse space directly.\footnote{In addition to the combination processing, we can perform different tasks in the sparse space, e.g., filtration and transfer. We leave these tasks to our future work. } 
Additionally, the proposed transformations and  combination method can be paralleled to enable efficient computation.



In summary, this paper makes the following contributions:

$\bullet$ It proposes a novel semantic transformation method which effectively connects dense and sparse spaces.

$\bullet$ It proposes to use a new activation function to achieve sparseness, which, to the best of our knowledge, has not been used before. The function works very well.

$\bullet$ It proposes a combination method that can encode sentence in the sparse space directly.

$\bullet$ The proposed methods have been evaluated using text classification and natural language inference tasks with promising results. Since the proposed transformations avoid large scale matrix multiplications in the combination procedure, it is also efficient. 


\section{Semantic Transformation}\label{sec:ST}
In this section, we first briefly describe the composition of Semantic Transformation (ST), and then elaborate on each component.
The proposed ST has three operations:\\

\begin{enumerate}
\item \textbf{SFT} (Semantic Forward Transformation). It takes a dense representation as input and transforms it into a higher dimensional sparse space.
\item \textbf{SBT} (Semantic Backward Transformation). It is the inverse of SFT, transforming representations from the sparse space back to the dense space. 
\item \textbf{SCSS} (Semantic Combination in the Sparse Space). It computes the sentence representation using its component word representations in the sparse space.\footnote{Note that although many operations can be done in the sparse space, the purpose of this paper is not to investigate all those operations. This paper mainly focuses on SCSS, the most basic operation in the sparse space for NLP.}
\end{enumerate}


\subsection{Semantic Forward Transformation} \label{sec:SFT}
SFT aims to discover the latent semantic aspects in a dense representation of word $\mathbf{x}$ and put them in a higher dimensional sparse representation $\mathbf{y}$. We assume $M$ is the number of latent semantic aspects\footnote{We set a limited number of semantemes because those latent semantemes are not all the semantemes in the real world but are the bases for composing real world semantemes.}, and each latent semantic aspect is represented by a vector, i.e. $\mathbf{b}_m \in R^{d}$ for the $m^{th}$ base. We define all the latent semantic aspects as the bases of semantemes in the real world, denoted by $\mathbf{B} = \{\mathbf{b}_1,\dots, \mathbf{b}_M\} \in R^{d \times M}$. Given $\mathbf{B}$, the function of SFT is to estimate the semantic distribution of the given dense representation over $\mathbf{B}$.

\vspace{+2mm}
\noindent \textbf{Definition  ($\mathbf{y}$): }
We define $-1 \prec \mathbf{y} \prec 1$, meaning that each element of $\mathbf{y}$ has a value in $(-1, 1)$. 

\vspace{+2mm}

The reasons for giving positive and negative values to elements in a sparse representation are that 1) negative values can represent ``negative semantemes''; 2) we can eliminate some meanings of elements (positive values) through simple operations between words, i.e., adding. Note that a negative value representing ``negative semantics'' of a given aspect does not mean that two words with opposite meanings have exactly corresponding positive and negative sparse representations. In the sparse space, we use the composition of semantemes to denote word meanings. This is in line with the human way of using words, e.g., the meaning of "not bad" can be obtained by adding the sparse representation of ``not'' to the sparse representation of ``bad''. In this sense, the meaning of ``not bad'' is a composition of several semantemes. 

\vspace{+2mm}
\noindent \textbf{Formulation of SFT: }
We adopt a multilayer perceptron (MLP)\footnote{Our approach is not limited to using multilayer perceptron (MLP). Other techniques, e.g., CNN may also be used.} integrated with the base $\mathbf B$ to build a SFT to perform its function. We first use a MLP $f(\cdot)$ to learn deep features of the dense representation $x$, and then use the features to compute the sparse distribution over the semantic bases. Formally, the $i^{th}$ layer in $f(\cdot)$ can be written as:
\begin{equation}\label{eq:ns1}
\begin{aligned}
    \mathbf{\mathbf{p}_i} = f_i(\mathbf{p}_{i-1}, \mu) = \sigma(\mathbf{w}_i\mathbf{p}_{i-1})
\end{aligned}
\end{equation}
where $\sigma$ is the activation function, and $\mathbf{w}_i$ is the parameter of the $i^{th}$ layer denoted by $\mu$; $\mathbf{p}_{i-1}$ is the output of ${(i-1)}^{th}$ layer and $\mathbf{p}_{0} = \mathbf{x}$. We denote the output of the last layer of $f(\cdot)$ as $\mathbf{p}$ and then integrate it with $\mathbf{B}$. The distribution over semantic bases can be computed by:
\begin{equation}\label{eq:ns2}
\begin{aligned}
    \mathbf{y} = S(\mathbf{p \cdot w}_f\mathbf{B})
\end{aligned}
\end{equation}
where $\mathbf{w}_f$ is a trainable parameter; $S( \cdot )$ is a specially designed activation function used to control the sparseness of the semantic distribution (discussed later). To sum up, SFT can be written as:

\begin{equation}\label{eq:ns3}
\begin{aligned}
    \mathbf{y} = \mathcal{SFT}(\mathbf{x}) = S(f(\mathbf{x}) \cdot \mathbf{w}_f\mathbf{B})
\end{aligned}
\end{equation}

\vspace{+2mm}
\noindent \textbf{Sparse Activation:}
Sparsity is enforced through penalties in most exist studies, such as $\ell_1$ regularizer~\cite{sun2016sparse}, average sparsity penalty~\cite{ng2011sparse}, and partial sparsity penalty~\cite{subramanian2018spine}. We call those methods \textit{penalty enforcing methods} which push the sparse representation close to either 0 or 1. 

\begin{figure}[t]
\centering 
\subfigure[]
{ 
\label{fig:activationfunction}
\includegraphics[width=0.6\columnwidth]{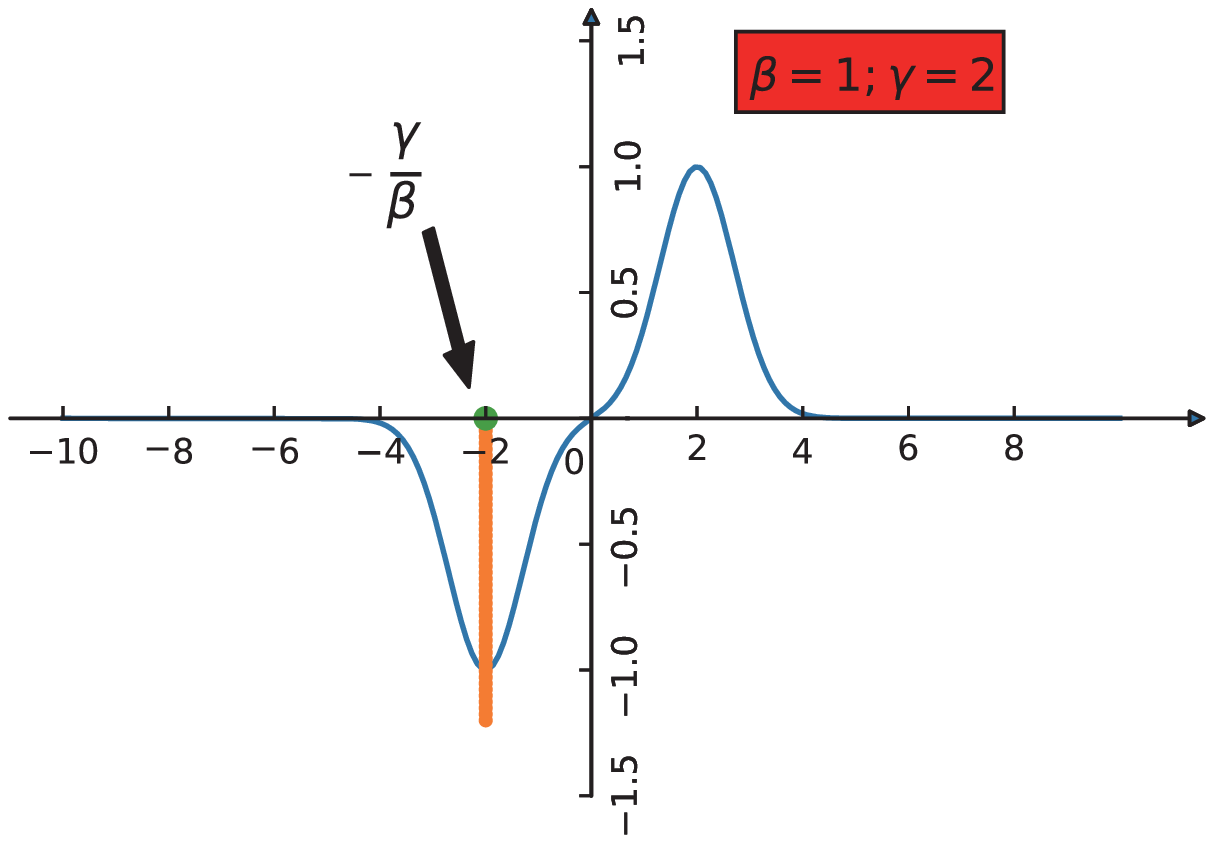}
} 
\subfigure[]
{ 
\label{fig:activationfunction2} 
\includegraphics[width=0.6\columnwidth]{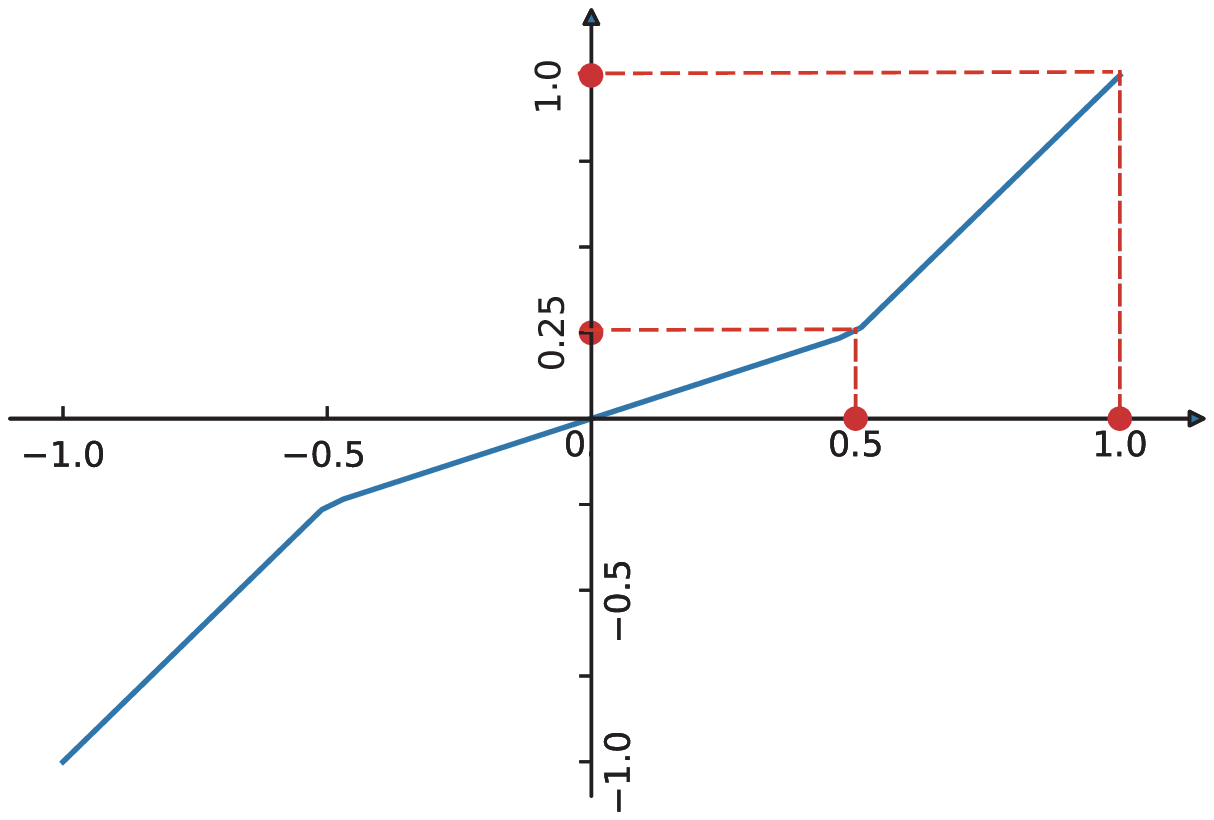}
} 
\caption{Activation curve.} 
\label{fig:attn_hm} 
\end{figure}

However, such penalties suffer from the initialization sensitivity problem as the penalties contain an initial interface which influences the distribution of the learned sparse representation significantly. To overcome the problem, we propose to use an activation function instead. We first give the formulation of the proposed activation function $S(\cdot)$ and then show its activation curve in Figure~\ref{fig:activationfunction}.
\begin{equation}\label{eq:ns4}
\begin{aligned}
    S(x) = e^{-(\beta x - \gamma)^2} - e^{-(\beta x + \gamma)^2}
\end{aligned}
\end{equation}
where $\beta$ and $\gamma$ are two hyper-parameters controlling the sparsity of the output. We set $\beta = 1$ and $\gamma = 2$ in our experiments.

Clearly, from Figure~\ref{fig:activationfunction}, we can see a large range of inputs of $S(\cdot)$ is mapped to 0, while the positions around $\pm \gamma/\beta$ will get high responses. Integrated with an objective loss function (depends on specific types of tasks, e.g., cross entropy for classification), SFT learns to give the relevant aspects/semantemes with predictions around $\pm \gamma/\beta$. In the case under the action of this activation function, we can learn sparse representations through the original objective function, not relying on enforcing penalties. Based on the experimental results, we will see that this activation function works very well on many 
datasets.


$S(\cdot)$ is non-linear and differentiable and its derivatives can be written as:
\begin{equation}\label{eq:ns5}
\begin{aligned}
    S'(x) = (-2 \beta^2 x + 2\gamma \beta \cdot \text{Sign}(x)) \cdot S(x)
\end{aligned}
\end{equation}
where $\text{Sign}(\cdot)$ is Sign function, and $\text{Sign}(0) = 0$. Clearly, the derivative of $S(\cdot)$ is easy to compute.

\subsection{Semantic Backward Transformation}
SBT is the inverse transformation of SFT, which transforms a sparse representation back to a dense representation. A straightforward way to achieve SBT is to use the sparse representation to do a weighted sum over the base $\mathbf{B}$. To increase the fitting ability of SBT, similar to SFT, we adopt a MLP $F(\cdot)$ to learn a deep dense representation. We formulate SBT as follows:
\begin{equation}\label{eq:ns6}
\begin{aligned}
   \mathbf{x} = \mathcal{SBT}(\mathbf{y}) =  F(\tanh( \mathbf{w}_b \cdot \mathbf{By}^T))
\end{aligned}
\end{equation}
where $\tanh$ is the Tanh activation function, and $\mathbf{w}_b$ is a trainable parameter. $F(\cdot)$ is a MLP with its own trainable parameters.

\subsection{Semantic Combination in Sparse Space}
This section proposes a \emph{Semantic Elimination (SE)} method to complete \textit{semantic combination} in the sparse space.\footnote{A sparse representation usually has a large number of dimensions (or aspects) but only a small number of dimensions have none zero values. Inherently, it is inappropriate to combine words' sparse representations into sparse sentence representations by using complex matrix transformation.} The main idea of SE is to use the negative values in the representation of one word to eliminate another word's semantics. That is also one of the reasons for defining negative values in the sparse representation. 
In this scenario, the sparse representation has two functions: (1) using positive values (positive semantemes) to denote which semantic meanings a word has and (2) using negative values (negative semantemes) to eliminate the semantemes that should not be present in the word. Below, we detail SE.

Due to the fact that a word's semantemes usually change with the nearby words or just the preceding word in a sentence, given a sentence, we propose to use the $i^{th}$ word's negative values to eliminate the $(i+1)^{th}$ word's positive values (semantemes). We call this elimination method \textit{Preceding Elimination (PE)}. After that, a nonlinear activation function must be followed to avoid the overall operation as a linear operation. Note, the activation function must go through the origin $(0, 0)$ in order to ensure the balance of positive and negative values. In this case, we specially designed an activation function, which we will elaborate it shortly. Then we add the sparse representations of all words in the sentence together after PE as the final sentence sparse representation.\footnote{In the sum vector, if an element is greater than 1 or less than -1, we reduce its absolute value to 1 without sign change.}


We designed an activation function, called `leaky' (its curve is shown in Figure~\ref{fig:activationfunction2}) to (1) decrease the small values of a sparse representation in order to prevent the system from producing new semantemes that shouldn't exist;
(2) make the SE sensitive to word order (in order to consider the information of word order) since the activation function is non-linear which enables non-commutativity of the whole SE over linear and non-linear operations. Note that `leaky' is used on sparse representations of words after preceding elimination. SE is formulated as:
\begin{equation}\label{eq:ns71}
\begin{aligned}
   \mathbf{s}_t = \sum^{t}_{i=1}\text{leaky}( -\text{Relu}(-\mathbf{y}_{i-1}) + \text{Relu}(\mathbf{y}_i) )
 \end{aligned}
\end{equation}
where $\mathbf{s}_t$ is the sparse representation of a sub-sentence from 1 to position $t$ produced by semantic combination in the sparse space. In this case, $\mathbf{s}_T$ denotes the sparse representation of a sentence with length $T$.

\subsection{Objective Function}
Overall, given a batch of data $\mathcal{D}$, our model is trained to minimize the following objective function:
\begin{equation}\label{eq:ns81}
\begin{aligned}
   \text{min}\ \ L(\mathcal{D}) = \text{PL}(\mathcal{D}) + \text{ML}(\mathcal{D}) + \text{BL}(\mathcal{D}) + \text{RL}^o(\mathcal{D})
\end{aligned}
\end{equation}
where $\text{PL}(\mathcal{D})$ denotes the prediction loss over the dataset, it depends on the task that the model is applied to; $\text{ML}(\mathcal{D})$ denotes the margin loss, it is performed to enlarge the margin of distances between sparse representations with different meanings; $\text{BL}(\mathcal{D})$ is a regularization used to constrain the norm of bases;   $\text{RL}^o(\mathcal{D})$ denotes the reconstruction loss, which is used to do model simulation (see below) and therefore \emph{it is optional}. Note, when applying our method only $\text{PL}(\mathcal{D})$ is necessary, $\text{ML}(\mathcal{D})$ and $\text{BL}(\mathcal{D})$ can be used to improve the model's performance. Next, we discuss these loss functions.

\vspace{+2mm}
\noindent \textbf{Prediction Loss (PL):} PL($\mathcal{D}$) is the training loss of the application task. For example, in our case, this loss is Cross Entropy for supervised classification.

\vspace{+2mm}
\noindent \textbf{Margin Loss (ML):} ML($\mathcal{D}$) is designed to enlarge the margin of distances between sparse representations with different meanings. We need ML to help training because we found that the margin of the learned sparse representation by optimizing PL is not clear or significant for separating positive and negative semantemes, which is undesirable for explanation. We then explore a new method for clear sparse representation learning, called Margin Loss, which makes the sparse representations having different meanings far from each other. 

In the scenario of classification, we leverage the class labels as supervising information to group the samples in a batch into each class, and then average the sparse representations of the instances in each class to represent the class. Formally, we assume $\mathbf{y}_{ci}$ is the averaged representation of the $i^{th}$ class. Then, based on the cosine similarity\footnote{Note that $\mathbf{y}_{ci}$ is not a sparse representation as it is the average of many sparse representations. Cosine similarity is appropriate for Margin Loss.}, we define ML($\mathcal{D}$) as follows:
\begin{equation}\label{eq:ns91}
\begin{aligned}
   \text{min} \ \ \text{ML}(\mathcal{D}) = \text{sum}(\mathbf{W} \odot ( \mathbf{Y}_c^T *\mathbf{Y}_c))
\end{aligned}
\end{equation}
where $\mathbf{Y}_c = \{\mathbf{y}_1,\dots,\mathbf{y}_N\}$, $N$ is the number of classes. $\odot$ denotes Hadamard product. $\mathbf{W} \in \mathbb{R}^{N \times N}$ is hyper-parameter used to control the updating direction and degree. $\mathbf{W}_{ij}$ is set to -1 if $i=j$, or 1 otherwise. This ensures a large margin between different classes by minimizing their inner product. Note that in some scenarios, especially sentiment classification, the distance of different classes belonging to the same positive (or negative) sentiment (e.g., strong and weak positive/negative classes) should not be enlarged much. In this case, we develop an exponential decay function to intuitively set $\mathbf{W}$:
\begin{equation} \label{eq:ns10}
   \mathbf{W}_{ij} = \begin{cases}
  \frac{1}{2}^{({N-1 - |i-j|})/{\tau}}, & \text{if} \ i \neq j \\
  -1, & \text{otherwise}
\end{cases}
\end{equation}
where $\tau$ is half-life, we set it to $(N-1)/2$.

\vspace{+2mm}
\noindent \textbf{Base Regularization: (BL):} Recall in the proposed semantic forward transformation method, base collection $\mathbf{B}$ is the key for obtaining the semantic distribution (semantic representation) of the given dense representation. Clearly, it is a projection procedure. Here, we argue that a larger projection will not ensure a better prediction. That is because representations with a large norm usually get a large projection, which is a point that conventional prediction methods ignore. The proposed Sparse Activation method eliminates this problem by giving large projections small responses. Similarly, inconsistent length of bases in $\mathbf{B}$ will cause different output (response) priors. To tackle this problem, we propose a base regularization to constrain the length of bases in $\mathbf{B}$ to equal to 1. Formally, BL is formulated as:
\begin{equation}\label{eq:ns11}
\begin{aligned}
   \text{min} \ \ \text{BL} = \sum_{m=1}^{M}(||\mathbf{b}_m|| - 1)^2
\end{aligned}
\end{equation}
where $\mathbf{b}_m$ is the $m^{th}$ base in $\mathbf{B}$. 

\vspace{+2mm}
\noindent \textbf{Reconstruction Loss (RL$^o$):} The proposed ST can easily do transformations among dense and sparse spaces, and learn sentence representation in the sparse space. In this case, ST could provide a sentence with both dense and sparse representations. One question that may be asked is whether the dense representations produced by ST through back transform can be used in place of dense representations directly learned by models in the dense space, e.g., LSTM? In this case, we propose reconstruction loss to minimize the construction error between the outputs of ST and LSTM. Another purpose of RL$^o$($\mathcal{D}$) is to control the meanings of the same word or sentence/phrase in different spaces to maintain consistency with the representations of a sentence and its phrases produced by LSTM as $\mathbf{X}$, then 
\begin{equation}
\begin{aligned}
  \text{RL}(\mathcal{D}) = \sum_{\mathcal{D}}\sum^{T}_{i=1}&( ||\mathbf{x}_{i}-\mathbf{x}_{i}'||_2^2 \\
  &+ ||\mathbf{s}_{i}-\mathbf{s}_{i}'||_2^2 + ||\mathbf{X}_{i}-\mathbf{X}_{i}'||_2^2 )
\end{aligned}
\end{equation}
 where $\mathbf{x}_{i}'=\mathcal{SBT}(\mathbf{y}_{i}),\mathbf{X}_{i}'=\mathcal{SBT}(\mathbf{s}_{i}),\mathbf{s}_{i}'=\mathcal{SFT}(\mathbf{X}_{i})$; $T$ is the length of the sentence. $\mathbf{x},\mathbf{y},\mathbf{s}_i$ have the same meanings as we defined before. $\mathbf{X}_i'$ denotes the dense representation constructed from sparse representation $\mathbf{s}_{i}$.
 This loss helps transform the representations in one space to another space while maintaining the semantic information consistency. The last term helps learn similar representations with LSTM.

\begin{table*}[!htbp]
\centering
\small
\caption{Average accuracy over all tasks. Y and X' are representations for making predictions (X' is the back transformation of Y; Y is the sparse representation). Helper loss refers to ML or BL. Note that only the experiments using X' as the representations for prediction has RL$^o$. RL$^o$ is not used when using Y as the prediction feature.}

\renewcommand\arraystretch{1.2}
\begin{tabular}{c|ccccccc}
\hline
Model&SNLI&MR&SST1&SST2&TREC\\
\hline
CNN \cite{kim2014convolutional} &59.71&76.10&36.80&{80.60}&\bf{90.20}\\
Transformer \cite{vaswani2017attention} &55.32&75.23&34.80&78.30&81.56\\
Capsule \cite{zhao2018investigating}&54.53&72.57&36.44&77.02& 82.31\\
LSTM \cite{hochreiter1997long} &66.66&71.04&36.96&75.11&87.60\\
\hline
ST$^\P$[X'] (without sparse activation or helper loss) & 32.90 & 61.07 & 29.97 & 68.04 & 63.40\\
ST$^\ddag$[X'] (with sparse activation, without helper loss) & 63.34 & 70.38 & 35.79 & 75.11 & 80.00 \\
ST$^\dag$[X'] (using the traditional penalty, without sparse activation or helper loss) & 59.89 & 65.51 & 33.97 & 65.25 & 73.40\\
ST[X'] (full model)& 66.58 & 71.16 & 38.69 & 76.03 & 87.06\\
\hline
ST$^\P$[Y] (without sparse activation or helper loss) & 62.46 & 68.32 & 35.60 & 71.33& 79.60\\
ST$^\ddag$[Y] (with sparse activation, without helper loss) & 63.53 & 76.38 & 38.24 & 78.33  & 86.20 \\
ST$^\dag$[Y] (using the traditional penalty, without sparse activation or helper loss) & 62.62 & 69.17 & 35.42& 71.33& 81.60\\
ST[Y] (full model)&\bf{66.85}&\bf{77.15}&\bf{41.78}&\bf{80.70}&87.80 \\
\hline
\end{tabular}
\label{tab:mainresult}
\end{table*}

\section{Experiments}

We evaluate the proposed method using one natural language inference dataset and four text classification datasets. The tasks act as good quality checks for the learned representations. The code is implemented with Pytorch and can be found here \footnote{\href{https://github.com/morning-dews/ST}{https://github.com/morning-dews/ST}} The five datasets are SNLI, MR, SST1, SST2 and TREC, detailed training/dev/test splits are shown on Table \ref{tab:data}:

$\bullet$ SNLI~\cite{bowman2015large}: a collection of human-written English sentence pairs manually labeled for balanced classification with the labels: entailment, contradiction, and neutral. This is the natural language inference dataset, which is also solved via classification. 

$\bullet$ MR v1.0\footnote{https://www.cs.cornell.edu/people/pabo/movie-review-data/}: Movie reviews with one sentence per review labeled positive or negative for sentiment classification.

$\bullet$ SST1\footnote{http://nlp.stanford.edu/sentiment/}: an extension of MR but with fine-grained labels: very positive, positive, neutral, negative, very negative. 

$\bullet$ SST2\footnote{http://nlp.stanford.edu/sentiment/}: same as SST1 but with neutral reviews removed and only using positive and negative labels. 

$\bullet$ TREC\footnote{https://cogcomp.seas.upenn.edu/Data/QA/QC/}: question samples that classify each question into one of 6 question types: about person, location, numeric information, etc. 

\begin{table}[!t]
\centering
\small
\renewcommand\arraystretch{1.2}
\caption{Summary statistics for the datasets after tokenization. c denotes the number of target classes.}
\begin{tabular}{ccccc}
\hline
Data&c&Train&Dev&Test\\
\hline
SNLI &3&549367&9842&9842\\
MR&2&8529&1067&1066\\
SST1&5&8544&1101&2210\\
SST2&2&6920&872&1821 \\
TREC&6&5452&500&500 \\
\hline
\end{tabular}
\label{tab:data}
\end{table}

\vspace{+2mm}
\noindent \textbf{Baseline:} Four widely used methods are employed as the baselines: 

(1) a 1-layer LSTM \cite{hochreiter1997long} with 300 hidden units; 

(2) a 3-layer Transformer \cite{vaswani2017attention} with 300 hidden units; 

(3) CNN \cite{kim2014convolutional}: We use exactly the same settings as the paper; 

(4) Capsule Network \cite{zhao2018investigating}. We adopted the code released by the authors and used trainable embeddings.

For our model, we adopt a MLP with 1 hidden layer (300 units) for forward transform and a MLP with 2 hidden layers (300 units) for backward transform. We set the length of semantic base to 1000.

\vspace{+2mm}
\noindent \textbf{Training details:} We adopt uniform settings for all baselines and our model: 

1) Adam optimizer for parameter updating with learning rate of 1e-4; trainable embeddings with size 300.

2) A MLP with 1 hidden layer as the classifier. For a fair comparison, the hidden unit size is set to 300 for LSTM, CNN, Transformer and Capsule. For our model, it is set to 64 when we use sparse representation to do the prediction and still 300 when we use back transformation representations as the prediction features.\footnote{In detail, the number of parameter of the classifiers for baselines and our model using back transformation representations is 300*300=90,000; while the number for our model using sparse representation is 1000*64=64,000.}

3) SNLI is the task of identifying the relationships between two given sentences. For each model, we first use it to encode the two sentences into the resulting representations respectively, and then concatenate the two sentence representations for the final prediction.

4) We report the average accuracy over 10 runs of the experiment on the test data. For each run, the maximum accuracy before early stopping is selected as the result of the current run.

\subsection{Results and Analysis}
Table \ref{tab:mainresult} shows the prediction accuracy of our model and the baselines. Table \ref{tab:time} gives the prediction run time. 
From Table \ref{tab:mainresult} and \ref{tab:time}, we can make the following observations:

\vspace{+1mm}
\noindent $\bullet$ The proposed Semantic Transform (ST) approach significantly outperforms LSTM on three datasets: SST1, SST2 and MR, and get comparable results with LSTM on SNLI and TREC. ST also markedly outperforms Transformer and Capsule on all five datasets, and outperforms CNN on four out of five datasets. Therefore, we can draw the conclusions that ST is an effective method to learn sentence representations in both dense and sparse spaces. 

\vspace{+1mm}
\noindent $\bullet$ ST$^\dag$ (including ST$^\dag$[X'] and ST$^\dag$[Y]) performs much worse than the proposed sparse activation method, which indicates the effectiveness of the proposed method. ST$^\ddag$ (including ST$^\ddag$[X'] and ST$^\ddag$[Y]) shows the proposed sparse activation plays an important role in our system, and it's very effective. And we will show that the proposed sparse activation method can ensure good sparseness of the representation through the analysis below.
The relatively worse results of ST$^\ddag$ (including ST$^\ddag$[X'] and ST$^\ddag$[Y]) also confirmed the effectiveness of helper losses.

\vspace{+1mm}
\noindent $\bullet$ In terms of efficiency, Table \ref{tab:time} shows that ST is 2-3 times faster than LSTM. 
ST is also markedly faster than Capsule and Transformer on all datasets. CNN is known as the fastest model and our method achieves comparable speeds with CNN.

In summary, considering that our work is only the first attempt, it performs quite well compared with highly researched and optimized LSTM, CNN, Capsule and Transformer models. We foresee that future work will significantly optimize our method. 

\begin{table}[!t]
\centering
\small
\renewcommand\arraystretch{1.2}
\caption{Average running time over all test sets (Minute) }
\begin{tabular}{ccccccc}
\hline
Model&SNLI&MR&SST1&SST2&TREC\\
\hline
CNN & 1.190&0.108&0.083 & 0.079& 0.035 \\
Transformer &1.810&{0.151}&0.140&{0.137}&0.041\\
Capsule&3.590&0.303&0.220&0.206&0.057\\
LSTM&2.096&0.186&0.168&0.142&0.049\\
ST&1.404&0.093&0.088&0.071&0.025 \\
\hline
\end{tabular}
\label{tab:time}
\end{table}

\begin{figure}[h]
\centering
\includegraphics[width=0.9\columnwidth]{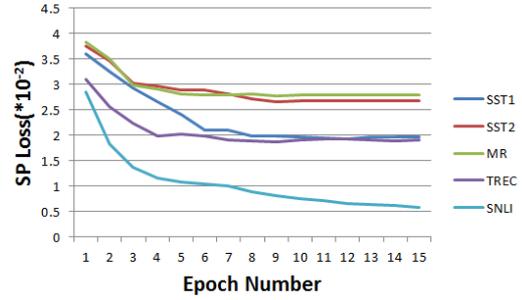}
\caption{Sparsity evaluation of sparse word representations (the legend is explained below)}
\label{fig:se}
\end{figure}

\vspace{+2mm}
\noindent \textbf{Sparsity Analysis:} Figure \ref{fig:se} shows the sparsity of the word sparse representations of the 5 datasets. Sparsity is evaluated using the following Sparse Evaluation (SE) function. We proposed this method because previous methods were not designed for sparse representations with both positive and negative values:
\begin{equation}
\begin{aligned}
   SE(\mathcal{D}) = \frac{1}{|\mathcal{D}|} \sum^{|\mathcal{D}|}_{i=1}(\sin(\pi \mathbf{y}_i))^2
   \end{aligned}
\end{equation}
As function $(sin(\pi y))^2$ has only three minimum points, -1, 0, 1, it is suitable for measuring the concentration degree of the components of sparse representations. 
Figure \ref{fig:se} shows a clear decline of SPLoss, which indicates a high concentration degree. Table \ref{tab:dsr} also gives the statistics about the distributions of the sparse representations. We can see that `zero' ($V<0.05$) takes a large portion of the sparse representations, which is desirable.
We can conclude that the learned sparse representations are indeed sparse.

\begin{table}[!htbp]
\centering
\small
\renewcommand\arraystretch{1.2}
\caption{Distribution of values in the sparse representations ($\%$). $V>0.6$ ($V<0.05$) shows the frequency of the values greater (less) than 0.6 (0.05)}
\begin{tabular}{ccccccc}
\hline
Metrics&SNLI&MR&SST1&SST2&TREC\\
\hline
$V>0.6$ &0.14&1.38&1.31&{1.71}&1.38\\
$V<0.05$&99.68&97.39&97.21&96.36&97.16\\
\hline
\end{tabular}
\label{tab:dsr}
\end{table}

\begin{figure}[h]
\centering
\includegraphics[width=0.9\columnwidth]{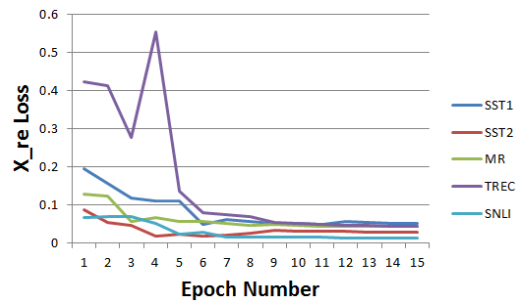}
\caption{Evaluation of the construction of $X$.}
\label{fig:ec}
\end{figure}

\begin{figure*}[h]
\centering
\includegraphics[width=1.95\columnwidth]{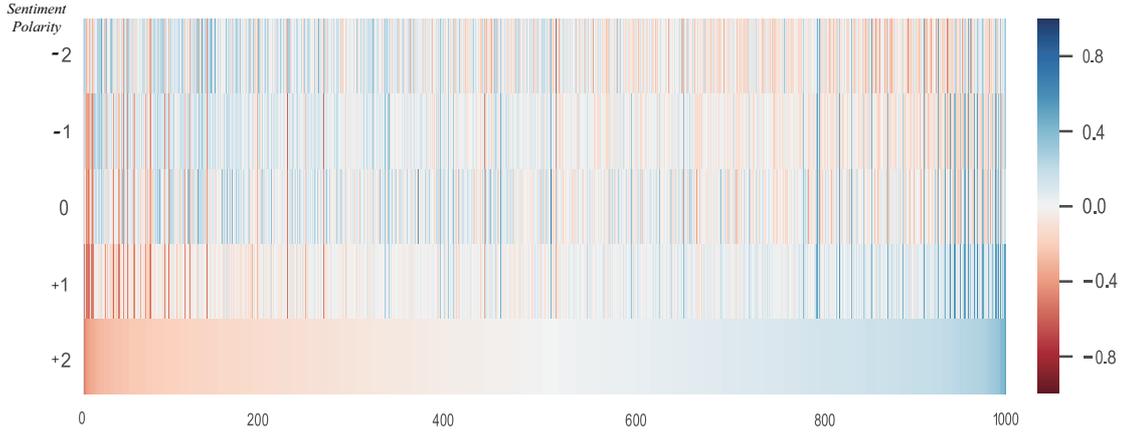}
\caption{Visualization of learned sparse representations.}
\label{fig:vsr}
\end{figure*}

\vspace{+2mm}
\noindent \textbf{Accuracy of Transformation:} We asked a question about the ability of ST to construct LSTM when we introduced the RL$^o$. Here, we analyze the transformation accuracy of the proposed method and give a positive answer to that question. From Table \ref{tab:mainresult}, we can see that ST[X'] achieves very similar results to those of LSTM. From the results, we can draw the conclusion that the dense representation generated by ST through backward transformation can achieve very similar results to those of LSTM. Further, we propose a measure to gauge the construction accuracy, named Construction Accuracy Metric (CAM), to evaluate the accuracy of transformation. CAM is formulated as the following function (results are shown in Figure \ref{fig:ec}):
\begin{equation}
\begin{small}
\begin{aligned}
   CAM(\mathcal{C}) = \frac{1}{J|\mathcal{C}|}\sum^{|\mathcal{C}|}_{i=1}\sum^{J}_{j=1}\frac{|X_{ij}-X_{ij}'|_2^2}{0.5*|X_{ij}|_2^2+0.5*|X_{ij}'|_2^2}
   \end{aligned}
  \end{small}
\end{equation}
where $X_{ij}$ is the original dense representation of a sub-sentence (generated by LSTM) and $X_{ij}'$ is the backward transformation result of its sparse representation; $\mathcal{C}$ denotes the test set, and $J$ is the length of the sentence. Clearly, this function can evaluate the similarity between $X$ and $X'$  as CAM will raise with the increasing of distance between $X$ and $X'$. Figure \ref{fig:ec} shows that the difference between $X$ and $X'$ is only about 5\%. Therefore, we can conclude that our model can construct the outputs of LSTM well.  

\vspace{+2mm}
\noindent \textbf{Interpretability Analysis:} Interpretability is one of the most desirable properties of sparse representations. Figure \ref{fig:vsr} shows the average sparse representation of five classes (tested on the test set of SST1) with different sentiment polarities (-2, -1, 0, 1, 2). Positive numbers refer to positive sentiment, and negative numbers refer to negative sentiment. In order to clearly visualize the differences in the learned representations over the five classes, we sort the bases based on the ascending order of the sparse representation values of +2 (very positive) class. 

From Figure \ref{fig:vsr}, we can see that there is a clear color difference for sentiment polarity class +2 and class -2. We can also see a similar phenomenon for sentiment polarity class +1 and class -1 but less pronounced as the their polarities are more similar. These observations demonstrate that the same bases obtain opposite values for classes of opposite sentiments. The bases generating distinct responses for classes with different sentiment polarities can be regarded as primary sentiment bases as they clearly indicate the semantic differences of the classes. In other words, the primary sentiment bases can be explained as sentiment bases. For example, the bases give positive response to positive classes but negative responses to negative classes are the positive sentiment bases, which directly indicate the sentiment polarities.

Comparing with positive and negative classes, neutral class shows relative mixed responses. That means neutral class has similar semantemes to those of both positive and negative classes. This demonstrates that the neutral class is more difficult to identify. 


\section{Related Work}

Sparse embeddings have been used in image \cite{ji2019partial,zhou2016sparse,zhang2016sparse}, signal \cite{caiafa2013computing,huang2007sparse}, and NLP \cite{subramanian2018spine,kober2016improving} applications.
 
Several sparse models have been proposed to produce sparse embeddings. For example, some previous works trained word embeddings with sparse or non-negative constraints~\cite{murphy2012learning,luo2015online}. Linguistically inspired dimensions \cite{faruqui2015sparse} is another way to increase sparsity and interpretability. SPINE (SParse Interpretable Neural Embeddings) \cite{subramanian2018spine}, a variant of denoising $k$-sparse autoencoder, can generate efficient and interpretable distributed word representations. Our method is different from these approaches. We not only construct sparse representations but also transform between dense and sparse spaces. We also combine word sparse representations to produce sentence representations. Some recent studies tried to achieve sparsity in novel ways \cite{park2017rotated}. We also proposed a novel method in this paper and experimentally verified its effectiveness. 

\section{Conclusion and Future Works}

This paper proposed a novel method to transform representations between dense and sparse spaces, and a technique to combine semantics in the sparse space. It also proposed and experimentally verified a new activation function that can be used to achieve sparseness. Natural language inference and text classification tasks were used to evaluate the proposed transformations with promising results. Based on this study, many other interesting directions can be pursued in the future, e.g.,

(1) As we discussed in the paper, the proposed method can construct the output of LSTM well. One future work is to apply ST to language modeling. In this case, the results can be used in many down stream tasks such as machine translation and dialogue systems. 

(2) With the help of ST, we can investigate the style transfer on similar tasks in the sparse space by direct semantic reversing. Also, we can use ST to filter out noises or undesirable information.

(3) Based on sparse representations, we can also explore semantic pattern recognition and transformation.

\bibliographystyle{acl_natbib}
\bibliography{acl2020}

\end{document}